\pdfoutput=1

\documentclass[11pt]{article}

\usepackage[final]{acl}

\usepackage{times}
\usepackage{latexsym}

\usepackage[T1]{fontenc}
\usepackage{xcolor}
\definecolor{MyPink}{RGB}{255, 20, 147}

\usepackage[utf8]{inputenc}

\usepackage{microtype}

\usepackage{inconsolata}

\usepackage{graphicx}

\usepackage{colortbl}
\usepackage{booktabs}
\usepackage{multirow}
\usepackage{tabularx}
\usepackage{amsmath}
\usepackage{amssymb}
\usepackage{mathtools}
\usepackage{xcolor}
\usepackage{tcolorbox}
\usepackage{hyperref}

\definecolor{headerbox}{HTML}{7BAFDE}
\definecolor{textbox}{HTML}{F5F5F5}
\definecolor{cblue}{HTML}{77AADD} %
\definecolor{cmint}{HTML}{44BB99} %
\definecolor{cgreen}{HTML}{AAAA00} %
\definecolor{corange}{HTML}{EE8866} %
\newcommand{\imagechain}{\textsc{ImageChain}}

\title{\imagechain: Advancing Sequential Image-to-Text Reasoning in Multimodal Large Language Models}

\author{
 \textbf{Danae S\'{a}nchez Villegas\textsuperscript{*}} \quad
 \textbf{Ingo Ziegler\textsuperscript{*}} \quad
 \textbf{Desmond Elliott}
\\
\\
University of Copenhagen, Department of Computer Science\\
\textsuperscript{*}Equal contribution\\
\texttt{\{davi,inzi,de\}@di.ku.dk}
}

\begin{document}
\maketitle

\begin{abstract}
Reasoning over sequences of images remains a challenge for multimodal large language models (MLLMs).
While recent models incorporate multi-image data during pre-training, they still struggle to recognize sequential structures, often treating images independently. This work introduces \imagechain{}, a framework that enhances MLLMs with sequential reasoning capabilities over image data by modeling visual sequences as a multi-turn conversation. In \imagechain{}, images are interleaved with corresponding textual descriptions to form a controlled dialogue that explicitly captures temporal dependencies and narrative progression. Our method optimizes for the task of \emph{next-scene description}, where the model generates a context-aware description of an upcoming scene based on preceding visual and textual cues. We demonstrate that our approach improves performance on the \emph{next-scene description} task -- achieving an average improvement from 3.7\% to 19\% in SimRate, a metric that quantifies semantic similarity to human-annotated ground truths. Moreover, \imagechain{} achieves robust zero-shot out-of-domain performance in applications ranging from comics to robotics. Extensive experiments validate that instruction-tuning in a multimodal, multi-turn conversation design is key to bridging the gap between static image understanding and temporally-aware reasoning.\footnote{Code, dataset, and checkpoints are publicly available at \url{https://github.com/danaesavi/ImageChain}.}
\end{abstract}

\section{Introduction}
\label{sec:introduction}

Multimodal large language models (MLLMs) such as GPT-4V \cite{achiam2023gpt}, MM1 \cite{mckinzie2025mm1}, and LLaVA-NeXT \cite{liu2024llavanext} have demonstrated impressive reasoning capabilities by integrating text and image inputs, advancing the state of visual-language understanding \cite{zhang2024mm}.
Standard tasks like image captioning and visual question answering (VQA) have driven significant progress in recognizing objects, attributes, and their relationships within individual images~\citep{stefanini2022show,srivastava2021visual,narins2024validated,wang2022modern}.
However, %
many real-world applications such as storytelling~\citep{huang-etal-2016-visual,wang2020storytelling,liu-etal-2023-visual-storytelling}, event comprehension~\citep{lei-etal-2020-likely, cheng-etal-2024-event}, and robotics~\citep{o2024open}, demand a deeper understanding of temporal and narrative progression across sequences of images.

\begin{figure}
    \centering
    \includegraphics[width=\linewidth]{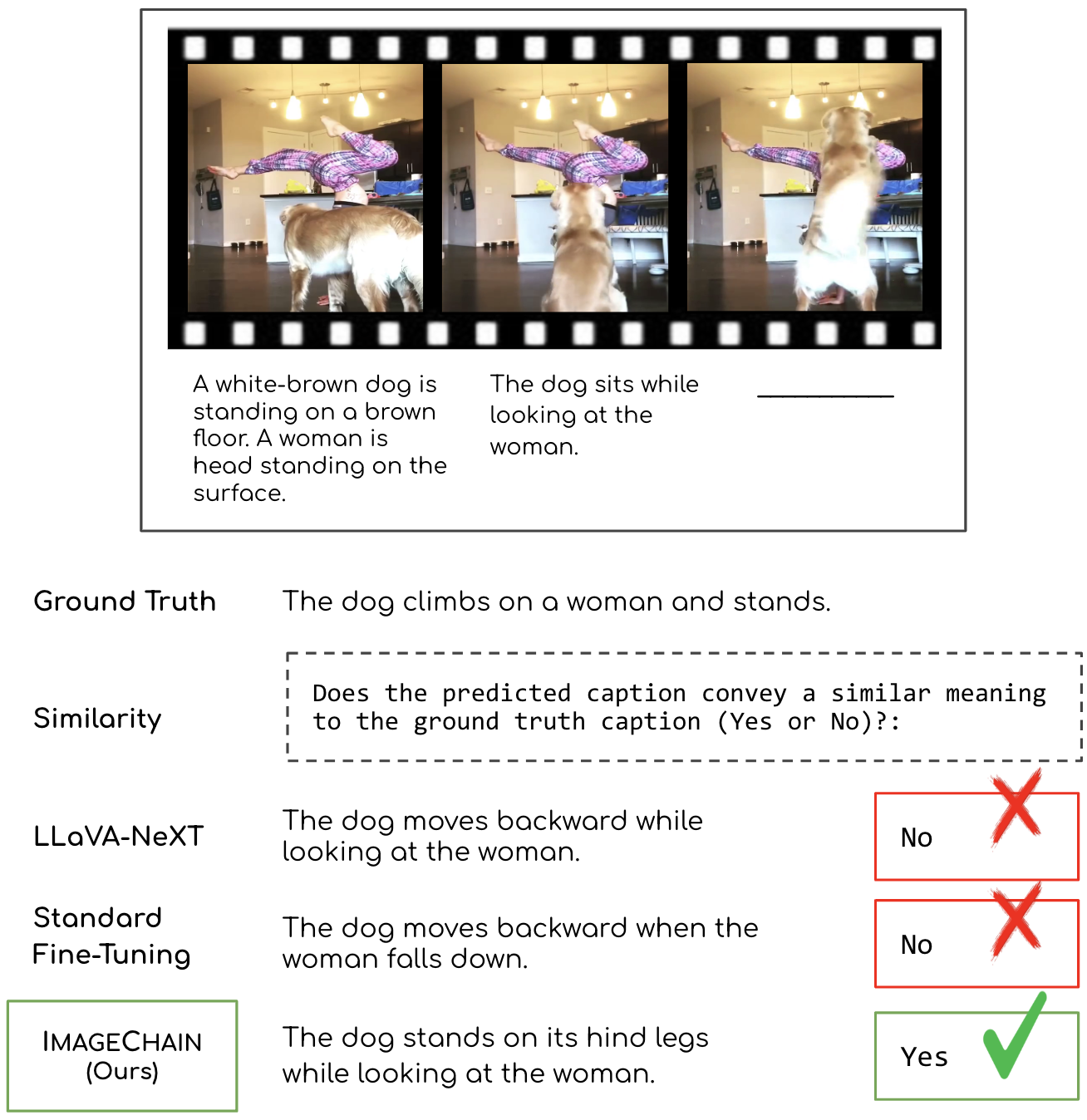}
    \caption[]{%
    Illustration of the proposed \emph{next-scene description} task. Models and techniques that process images independently, such as LLaVA-NeXT and standard fine-tuning, fail to capture the correct progression, leading to errors. In contrast, \imagechain{} explicitly models sequences as multi-turn conversations, enabling it to generate a more accurate description, closely aligning with the human-annotated ground truth.
    }
    \label{fig:example}
\end{figure}

Recent methods have extended MLLMs such as LLaVA-NeXT, Mantis~\citep{jiang2024mantis}, and Qwen2-VL~\citep{wang2024qwen2} to handle multi-image inputs. However, they typically process images independently or summarize entire scenes, rather than explicitly modeling the evolution of events over time.
In contrast, \emph{sequential image reasoning} requires a model to capture dependencies across frames to predict future actions. Figure~\ref{fig:example} illustrates our proposed task, \emph{next-scene description}, which consists of generating a text description of a visual scene based on a sequence of preceding frames and their corresponding descriptions -- a challenge that multimodal models, such as LLaVA-NeXT, have yet to overcome.

To address this gap, we introduce \imagechain{}, an efficient framework that provides MLLMs with explicit sequential reasoning capabilities.
By reformulating a visual sequence as a multi-turn conversation, \imagechain{} interleaves images with their corresponding text descriptions to build a sequential context for generating the \emph{next-scene description}.
Our approach achieves substantial improvements on similarity rate (SimRate), a metric that quantifies semantic similarity to human-annotated ground truths, across both in-domain and out-of-domain tasks using only approximately 4,000 training samples. To support our method, we repurpose StoryBench~\citep{bugliarello2023storybench}, a video dataset with human-annotated descriptions, to create and introduce \emph{StoryFrames} -- a high-quality, temporally coherent corpus tailored towards sequential image-text reasoning across different context lengths. StoryFrames provides annotated samples that enable \imagechain{} to efficiently adapt and learn robust temporal dependencies with minimal data. Our contributions are as follows:
\begin{itemize}
\item \textbf{Framework:} We introduce \imagechain{}, an image-to-text reasoning adaptation framework that models image sequences as multi-turn conversations for generating \emph{next-scene descriptions}, achieving an overall SimRate of 19\% versus 3.7\% for standard MLLMs.
\item \textbf{Robust Out-of-Domain Performance:} In robotics, \imagechain{} achieves an F1 score of 27.1—almost double the 14.4 of standard fine-tuning -- along with gains in structured settings such as comics.
\item \textbf{Context Length Ablations:} We show that training across multiple context lengths consistently outperforms training on single length, suggesting that exposure to varied temporal spans enhances sequential reasoning.
\item \textbf{StoryFrames:} We repurpose StoryBench, a video dataset with human-annotated descriptions, into StoryFrames, a corpus of 8,881 samples for facilitating research on general-purpose paired sequential vision and text data.
\end{itemize}

\begin{figure*}[t!]
    \centering
    \includegraphics[width=\linewidth]{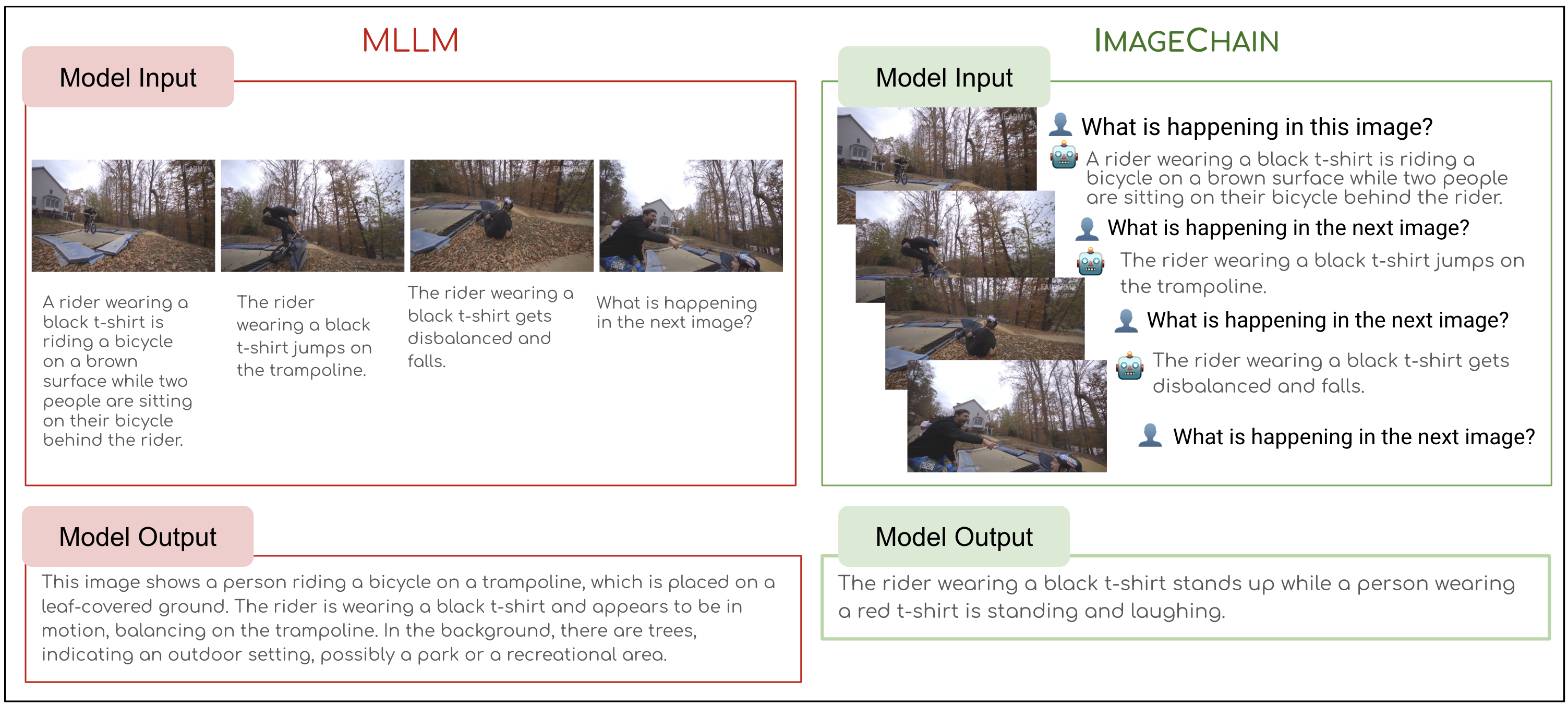}
    \caption[]{Comparison of multi-image sequential reasoning between a standard Multimodal Large Language Model (MLLM) and our proposed model \imagechain{}. The left side shows the output of MLLM, which fails to accurately describe the next event in the image sequence. The right side presents \imagechain{}, an image-to-text reasoning adaptation framework that models image sequences as multi-turn conversations, enabling a more accurate and temporally aware description of the next scene.}
    \label{fig:method}
\end{figure*}

\section{Related Work}
\label{sec:related-work}

\subsection{Evolution of Sequential Visual Reasoning}
Visual storytelling emerged as an early effort to generate coherent narratives from image sequences~\citep{huang-etal-2016-visual}.
Initial approaches relied on convolutional neural networks~\citep{lecun1995convolutional} for visual feature extraction paired with recurrent neural networks~\citep{hochreiter1997long} for narrative generation~\citep{gonzalez2018contextualize, kim2018glac}.
Although these methods demonstrated that visual and textual information could be integrated to produce compelling stories, they relied on overly broad scene summarization techniques~\citep{hong-etal-2020-diverse,wang2020storytelling}, rather than explicitly modeling the temporal dependencies required to predict future events.

With the introduction of MLLMs such as GPT-4V~\cite{achiam2023gpt}, MM1~\citep{mckinzie2025mm1}, and LLaVA-NeXT~\citep{liu2024llavanext}, significant progress has been made in static image understanding for tasks like image captioning and visual question answering~\citep{liu-etal-2023-visual-storytelling,lin2024improving, zhang2025let}.
Despite these successes, many MLLMs remain optimized for static or non-sequential multi-image inputs, limiting their ability to capture temporal dynamics.
This limitation motivates our work to develop a method that explicitly models temporal dependencies across image sequences.

\subsection{Challenges in Context Length and Generalization}
A critical challenge in sequential visual reasoning lies in effectively handling variable-length temporal contexts~\citep{zhou2024rethinkingvisualdependencylongcontext}.
Many existing models are trained on short or fixed-length sequences and thus struggle when presented with complex temporal spans or variable length contexts~\citep{thawakar2025llamavo1rethinkingstepbystepvisual}.
Furthermore, models show particular difficulties handling subtle temporal dependencies and structured event progressions where actions follow constrained logical sequences, such as in comics or robotics~\citep{wang-etal-2024-mementos}.
A recent analysis suggests that MLLMs rely on surface-level cues~\citep{zhou2024rethinkingvisualdependencylongcontext}, leading to performance degradation when processing extended contexts or adapting to novel scenarios~\citep{imam2025multimodalllmsvisualtemporal}.
These challenges underscore the need for approaches that capture variable-range temporal dependencies and generalize more robustly across diverse domains.

\subsection{Instruction Tuning Over Multi-Turn Conversations}
Instruction tuning~\citep{sanh2022multitask,wei2022finetuned} enhances LLMs by improving their ability to follow general task-agnostic directives while %
requiring limited training data~\citep{zhou2024lima}.
This methodology diverges from conventional fine-tuning approaches by exposing the model to varied instructional formulations, which ensures that the emphasis lies on adhering to directives rather than on task-specific details.~\citep{zhang2024instructiontuninglargelanguage}.
Empirical evaluations demonstrate that instruction-tuned variants consistently outperform competing baselines in output quality across open-ended~\citep{jha2023limit}, knowledge~\citep{jiang-etal-2024-instruction}, and reasoning~\citep{tangmathscale} tasks.

Instruction tuning over multi-turn conversations extend this approach by introducing sequential dependencies between interactions, where each turn builds on prior user-model exchanges~\citep{zhang2025surveymultiturninteractioncapabilities}.
Despite these advances in LLMs, such multi-turn conversational techniques have not yet been applied to multimodal settings.

\imagechain{} extends the multi-turn conversational paradigm to MLLMs for improved sequential visual reasoning by leveraging the strengths of instruction tuning and multi-turn interactions for integrating sequential visual data.
Rather than relying on emerging user-model interactions, we explicitly structure the conversation as a fixed sequence~\citep{wang-etal-2024-instruct}.
We interleave visual embedding tokens with scene descriptions to build a controlled context that emphasizes temporal dependencies.
Each turn poses a targeted question about an upcoming scene, with the expected response being a text description.
We call this task \emph{next-scene description} (Figure \ref{fig:method}), where the goal is to generate an accurate description of a visual scene based on preceding frames and annotations.

\begin{figure*}
\centering
\small

\begin{tcolorbox}[
    colback=textbox,
    colframe=headerbox,
    colbacktitle=headerbox,
]
    \texttt{\textcolor{cgreen}{<s>} \textcolor{cmint}{USER:}} What is happening in this image? \textcolor{corange}{\texttt{<Image><image></Image>}}\newline
    \texttt{\textcolor{cmint}{ASSISTANT:}} A rider wearing a black t-shirt is riding a bicycle on a brown surface while two people are sitting on their bicycle behind the rider. \texttt{\textcolor{cgreen}{</s>}}\newline
    \texttt{\textcolor{cgreen}{<s>} \textcolor{cmint}{USER:}} What is happening in the next image? \textcolor{corange}{\texttt{<Image><image></Image>}}\newline
    \texttt{\textcolor{cmint}{ASSISTANT:}} The rider wearing a black t-shirt jumps on the trampoline. \texttt{\textcolor{cgreen}{</s>}}\newline
    \texttt{\textcolor{cgreen}{<s>} \textcolor{cmint}{USER:}} What is happening in the next image? \textcolor{corange}{\texttt{<Image><image></Image>}}\newline
    \texttt{\textcolor{cmint}{ASSISTANT:}}  The rider wearing a black t-shirt gets disbalanced and falls. \texttt{\textcolor{cgreen}{</s>}}\newline
    \texttt{\textcolor{cgreen}{<s>} \textcolor{cmint}{USER:}} What is happening in the next image? \textcolor{corange}{\texttt{<Image><image></Image>}}\newline
    \texttt{\textcolor{cmint}{ASSISTANT:}}
\end{tcolorbox}
\caption{%
Multi-turn conversation design for a story with four scenes, where each turn corresponds to a scene. A turn begins with a user question and ends with the assistant's response. The context includes three completed turns (i.e., three scenes), along with the next user question and the corresponding visual cue, which are used to generate the \emph{next-scene description}.
}
\label{fig:multi-turn-conv}
\end{figure*}

\section{\imagechain: Optimizing MLLMs for Sequential Image-to-Text Reasoning}
\label{sec:method-data} 
\imagechain{} enhances an MLLM's ability to reason over sequential image data by optimizing the task of \emph{next-scene description}.
We assume a multimodal language model that can process both text and visual features.

\subsection{Problem Setup}
Let $S$ be a story represented as a sequence of scenes $S = \langle s_t \rangle_{t=1}^{T}$.
Each scene $s$ consists of a sequence of frames $V = \langle v_k \rangle_{k=1}^{K}$ and a textual description $D$ that corresponds to the entire scene, expressed as $s = (V, D)$. The number of scenes in a story and the number of frames per scene can vary across different stories.

\subsection{Obtaining Visual Scene Representations}
Each sequence of frames $V$ is transformed into a fixed-size representation to be paired with the sole description $D$.
Given a visual encoder $f_{\theta}(\cdot)$, each frame $v_k$ is mapped to a feature vector as $z_k = f_{\theta}(v_k)$, where $z_k \in \mathbb{R}^d$.

We then average the feature vectors $z_k$ over all frames to compute the scene-level visual representation as follows: $\,$
$\bar{V} = \frac{1}{K} \sum_{k=1}^{K} z_k.$ 
The averaged feature vector $\bar{V}$ serves as a fixed-size summary of the visual content in $V$, which is then paired with the corresponding textual description $D$.

\subsection{Multi-Turn Conversation Construction}
To capture the sequential nature of each story, we frame every scene as part of a multi-turn conversation between the user and the model.
For a given story $S$, let $\tau \in \{1, 2, \ldots, T\}$ denote the turns, where each turn $\tau$ is a triple $\left(Q_{\tau},\, \bar{V}_{\tau},\, D_{\tau}\right)$.
$Q_{\tau}$ is a predefined question string that asks from a user perspective ``\textit{What is happening in this image?}'' if $\tau = 1$ to start the conversation, or ``\textit{What is happening in the next image?}'' for all $\tau > 1$ to proceed to the next turn in the conversation.
Therefore, each story $S$ can be represented by a conversation context $C$ of $T$ turns:
\begin{equation}
C = \Bigl\langle
   (Q_{1},\, \bar{V}_{1},\, D_{1}), \,
   \ldots, \,
   (Q_{T},\, \bar{V}_{T},\, D_{T})
\Bigr\rangle.
\end{equation}

\subsection{Instruction Fine-Tuning Objective}
We fine-tune the model using standard supervised next-token prediction over the multi-turn conversation context $C$.
Let $W_C = \langle w_1, w_2, \ldots, w_N \rangle$ denote the concatenated sequence of all text tokens in the context $C$.
Our training goal for an individual story is to minimize the cross-entropy loss over all text tokens $w_i$ conditioned on all preceding text tokens and the visual context:
\begin{equation}
    \mathcal{L}_\text{IC} = -\sum_{i=1}^{N} \log p \Bigl(w_i \, \big| \, w_{1:i-1}, \bigl\{\bar{V}_\tau \bigr\}_{\tau=1}^T  \Bigr).
\end{equation}
The visual embeddings $\bar{V}$ are provided only for conditioning and do not contribute to the loss.
By enforcing a multi-turn structure that explicitly exposes the model to sequential context at each turn, minimizing $\mathcal{L}_{\text{IC}}$ improves the model's temporal reasoning over sequential image data.

\subsection{Next-Scene Description Generation}
To generate the \textit{next-scene description} $\hat{D}_\tau$ at turn $\tau$, we condition the instruction-tuned model on the \emph{completed} turns $\{1, 2, \ldots, \tau-1\}$, plus the current question $Q_\tau$ and the embedding $\bar{V}_{\tau}$.
Note that the next-scene description $D_\tau$ is withheld:
\begin{multline}
C_{\tau} = 
\Bigl\langle
(Q_{1},\, \bar{V}_{1},\, D_{1}), \,
\ldots, \\
(Q_{\tau-1},\, \bar{V}_{\tau-1},\, D_{\tau-1}), \,
(Q_{\tau},\, \bar{V}_{\tau},\, \square)
\Bigr\rangle.
\end{multline}
At inference time, the model auto-regressively generates the next-scene description $\hat{D}_\tau$ based on the provided multi-turn context $C_\tau$.
An example context with $\tau = 4$ turns is illustrated in Figure \ref{fig:multi-turn-conv}.

\section{The StoryFrames Dataset}
\label{sec:data}

We adapt the StoryBench dataset~\citep{bugliarello2023storybench} to create \emph{StoryFrames} -- a dataset specifically designed for the \emph{next-scene description} task on sequential visual reasoning.
StoryFrames repurposes human-annotated temporal segments from three distinct sources:
Oops~\citep{epstein2020oops}, which captures unexpected actions; 
DiDeMo~\citep{anne2017localizing}, which provides event grounding through natural language;
and UVO~\citep{wang2021unidentified}, which focuses on object-centric reasoning in video sequences.
In this dataset, each \emph{story} represents a single ``sample'' consisting of a sequence of scenes.
Every scene is defined by human-annotated start and end points, accompanied by a textual description covering its duration.
Each scene is further divided into multiple extracted frames, forming a hierarchical structure that supports detailed sequential reasoning.

\paragraph{Frame Extraction.}
To extract frames, we implement an adaptive frame sampling strategy that adjusts the number of frames based on the total duration of each story. 
For stories up to 5 seconds, we extract 8 frames; for 5 to 10 seconds, 12 frames; for 10 to 15 seconds, 15 frames; for 15 to 30 seconds, 20 frames; and for stories exceeding 30 seconds, 25 frames. 
This approach ensures broad temporal coverage while maintaining manageable computational requirements. 
Moreover, to preserve narrative continuity and prevent overlap between adjacent scenes, a 0.2-second temporal offset is introduced between consecutive action segments.

\paragraph{Frame Allocation Across Scenes.}
Within each story, frame allocation to individual scenes is performed proportionally to the duration of each scene, while ensuring that every scene receives at least two frames.
Formally, for a scene $s$ with duration \(m_s\) in a story $S$ of total duration $M$ and given a total frame budget $F$, the number of frames allocated is computed as follows:
$K_s = \max\left(2, \left\lfloor \frac{m_s}{M} \cdot F \right\rfloor\right).$

\begin{figure}
    \centering
    \includegraphics[width=0.95\linewidth]{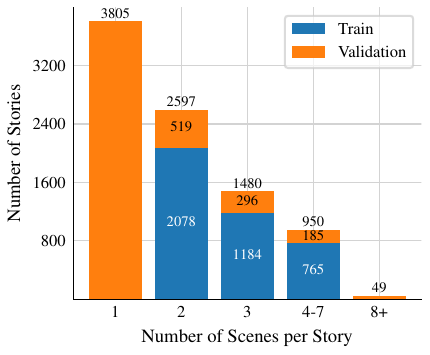}
    \vspace{-0.2cm}
    \caption{Distribution of scenes in our introduced StoryFrames dataset. A \emph{story} corresponds to one ``sample'', whereas multiple \emph{scenes} make up one story.}
    \label{fig:storyframes-dist}
\end{figure}

\paragraph{StoryFrames Organization.}
StoryFrames comprises 8,881 sequences and is organized into distinct subsets based on the number of scenes per story  (Figure \ref{fig:storyframes-dist}) to enable evaluation of models across varying context lengths and complexity.

\paragraph{Dataset Splits.}
For sequences containing 2 to 7 scenes, we maintain an 80-20 train-validation split, %
while single-scene sequences and those with 8 or more scenes are reserved for further validation to assess generalization to extreme cases. Figure~\ref{fig:storyframes-dist} visualizes the dataset distribution.

\begin{table*}[t]
\renewcommand{\arraystretch}{1.1}
\centering
\begin{tabularx}{\textwidth}{l *{5}{>{\centering\arraybackslash}X} *{2}{>{\centering\arraybackslash}X}}
\toprule
Model             & C2   & C3   & C4   & C5   & C6   & C4--6  & C2--6  \\
\midrule
MLLM              & 5.20  & 2.36  & 1.67  & 1.92  & 0.00    & 1.58    & 3.70    \\
MLLM-ICL          & 0.19  & 0.34  & 0.00    & 0.00    & 0.00    & 0.00    & 0.20    \\
MLLM-FT           & 18.11 & 19.59 & 15.83 & 3.85  & 15.38 & 12.13   & 17.52   \\
VisualContext     & 12.52 & 16.22 & 11.67 & 1.92  & 15.38 & 8.96   & 13.01   \\
FinalScene        & 9.63  & 10.47 & 11.67 & 1.92  & 7.69  & 8.44    & 9.71    \\
\imagechain{}-NoFT   & 6.94  & 6.08  & 12.50 & 0.00    & 15.38 & 8.96    & 7.11    \\
\imagechain{}        & \textbf{18.30} & \textbf{21.28} & \textbf{19.17} & \textbf{9.62} & \textbf{30.77} & \textbf{16.87} & \textbf{19.02} \\
\bottomrule
\end{tabularx}
\caption{Similarity Rate (SimRate, \%) across varying evaluation context lengths for \imagechain{} and baseline models trained on C2-7. \imagechain{} consistently outperforms all baselines and achieves the highest SimRate across all evaluated contexts.}
\label{tab:main-table}
\end{table*}

\section{Experimental Setup}
\subsection{Models}
We evaluate a series of model configurations ranging from unmodified baselines to fully fine-tuned variants. %
This systematic evaluation allows us to isolate the contributions of fine-tuning, visual context, and structured dialogue, thereby assessing the necessity and effectiveness of \imagechain{}.

\textbf{MLLM:} The baseline MLLM, designed to assess task performance without any task-specific modifications.

\textbf{MLLM-ICL:} An in-context learning baseline that receives three demonstrations of the task, designed to evaluate whether sequential image reasoning can be learned solely through seeing examples. %

\textbf{MLLM-FT:} MLLM with standard fine-tuning on the StoryFrames dataset, allowing a direct comparison with our proposed method.

\textbf{VisualContext:} A variant fine-tuned using only the visual context from preceding actions, i.e., all textual descriptions are omitted, enabling us to assess the contribution of text in the prediction task.

\textbf{FinalScene:} A variant that represents image captioning by fine-tuning the model to predict the next description based solely on the final visual action, designed to analyze whether incorporating context improves performance.

\textbf{\imagechain-NoFT:} Our approach that employs a multi-turn conversation structure but without any fine-tuning, isolating the effect of prompting with the structured dialogue format.

\textbf{\imagechain{}:} The complete method that integrates multi-turn prompting with fine-tuning on StoryFrames, explicitly modeling sequential dependencies to enhance \emph{next-scene descriptions}.

Our primary experiments are conducted using LLaVA-v1.6 with Vicuna-7B~\citep{vicuna2023} and CLIP~\citep{radford2021learning} variant~\citep{liu2024llavanext} as the base MLLM architecture. To assess the robustness of our findings and ensure that \imagechain{}'s improvements generalize across backbones, we replicate our main in-domain experiments (see Section~\ref{sec:experiments-datasets}) using Qwen2-VL-7B~\citep{wang2024qwen2}. For Qwen, we select the first frame per scene to avoid inconsistent representations from averaging, as its visual tokens are adaptively generated compared to LLaVA-1.6's static CLIP features. Unless otherwise specified, results presented are based on the LLaVA backbone. Training and implementation details are described in Appendix~\ref{sec:appendix-implementation_details}.

\subsection{Dataset and Context Length Splits}
\label{sec:experiments-datasets}
To evaluate the impact of temporal context on generating \emph{next-scene descriptions}, we use stories from the StoryFrames dataset with 2 to 7 scenes.
In our setup, the model is given a varying number of preceding scenes to predict the textual description for the current one.
This analysis reflects real-world applications that require reasoning over differing lengths of context. We define 3 evaluation settings:

\textbf{C2:} The model uses one preceding scene (a total of 2 scenes, including the current one).

\textbf{C3:} Here, the model uses two preceding scenes (totaling 3 scenes).  

\textbf{C4--7:} In this setting, the model receives a longer sequence, with three to six preceding scenes (corresponding to sequences of 4 to 7 scenes in total).

\textbf{C2--7:} The model is exposed to the full range of context lengths available (i.e., from 2 to 7 scenes). This range captures both short-term dependencies and more extended temporal structures and is designed to improve the model’s ability to generalize across varying context lengths.

By comparing model performance across the fixed settings (C2, C3, C4--7, C2--7), we are able to isolate the contribution of context length to \emph{next-scene description} and provide insights on optimizing for specific context lengths.

\subsection{Evaluation}
\paragraph{LLM-as-a-Judge.}
To quantitatively assess the quality of the generated descriptions, we use similarity rate (\textbf{SimRate}), an adapted version of win rate~\citep{chiang2024chatbot}, as a metric.
Given the inherent variability in how visual events can be described literally, conventional metrics relying on n-gram overlap are inadequate for capturing semantic equivalence~\citep{culy2003limits,bulian-etal-2022-tomayto}.
Instead, we follow recent literature~\citep{li2024llmsasjudgescomprehensivesurveyllmbased,ziegler2024craft,li2025generationjudgmentopportunitieschallenges} and adopt LLMs as evaluators~\citep{eldan2023tinystories}.
For our generated \emph{next-scene description} $\hat{D}_\tau$, Llama 3.1 70B~\citep{dubey2024llama} determines whether or not it conveys a similar meaning to the ground truth scene description $D_\tau$, effectively serving as a proxy for human evaluators~\citep{alpaca_eval,chiang2024chatbot}.
The prompt used for this framework is detailed in Appendix~\ref{sec:appendix-eval-prompt}.
The overall SimRate follows as the fraction of comparisons where model-generated descriptions are judged semantically similar to the human-annotated ground truth descriptions.
Section~\ref{sec:human-validation} presents a human validation study that confirms that our automated SimRate metric aligns well with human judgments.

\paragraph{Out-of-Domain Generalization.}
To further assess model generalization, we evaluate performance on several out-of-domain datasets that target diverse sequential reasoning challenges.
Specifically, we test on three datasets derived from the Mementos benchmark~\citep{wang-etal-2024-mementos}. \textbf{Comics}, which features wordless multi-panel comics;
\textbf{Daily-Life (DL)}~\citep{xiao2021next}, consisting of videos depicting everyday activities;
and \textbf{Robo}~\citep{o2024open}, which contains robotics tasks.
For evaluation, we follow Mementos and extract behavioral cues, (e.g., key verbs or verb phrases, from generated descriptions using GPT-4V~\citep{achiam2023gpt} and compared against human annotations using F1 score.

\begin{table}[t]
\centering
\renewcommand{\arraystretch}{1.07}
\setlength{\tabcolsep}{4pt}
\resizebox{\columnwidth}{!}{%
\begin{tabular}{lccccccc}  
\toprule
Model & C2 & C3 & C4 & C5 & C6 & C4--6 & C2--6 \\
\midrule
MLLM                 & 6.7 & 9.5 & 7.5 & 1.9 & 0.0 & 5.4 & 7.3 \\
MLLM-ICL             & 2.3 & 2.7 & 2.5 & 0.0 & 0.0 & 1.6 & 2.3 \\
MLLM-FT              & \textbf{20.0} & 21.3 & 16.7 & 11.5 & 23.1 & 15.7 & \textbf{19.6} \\
VisualContext        & 13.3 & 17.6 & 16.7 & 9.6 & 38.5 & 16.2 & 15.1 \\
FinalScene           & 14.1 & 17.6 & 17.5 & 11.5 & 30.8 & 16.8 & 15.6 \\
\imagechain{}-NoFT   & 9.3 & 17.9 & 15.8 & \textbf{13.5} & 38.5 & 16.8 & 13.2 \\
\imagechain{}        & 18.7 & \textbf{22.3} & \textbf{19.2} & 7.7 & \textbf{46.2} & \textbf{17.8} & \textbf{19.6} \\
\bottomrule
\end{tabular}%
}
\caption{%
SimRate (\%) on StoryFrames using Qwen2-VL as an alternative backbone. \imagechain{}'s sequential reasoning advantages generalize across architectures. As context length increases, our explicit multi-turn formulation yields progressively stronger scores compared to standard fine-tuning (MLLM-FT).}
\end{table}

\section{Results and Analysis}
\label{sec:results}

\subsection{In-Domain Performance}

\paragraph{Explicit Sequence Modeling Improves Temporal Reasoning.}
Table~\ref{tab:main-table} shows that our proposed \imagechain{} model consistently achieves the highest SimRate across all evaluated context lengths, demonstrating the value of explicitly modeling a sequence of images for \emph{next-scene description} task.
\imagechain{} also significantly outperforms standard fine-tuning (MLLM-FT), with a notable 15.39 percentage point improvement in SimRate on C6, the longest evaluated context length.
Additionally, \imagechain{} achieves 19.02\% on C2-6, surpassing VisualContext (13.01\%) and FinalScene (9.71\%), underscoring the value of incorporating both images and textual description history.
Even without fine-tuning, \imagechain{}-NoFT outperforms MLLM on C2-6 (7.11\% vs. 3.70\%), demonstrating the benefits of explicitly modeling image sequences. However, fine-tuning remains essential for optimal performance, suggesting that MLLMs lack strong inherent temporal reasoning skills without additional training.

\paragraph{\imagechain{} Generalizes Across Backbones.}
To evaluate the backbone robustness of \imagechain{}, we replicate our main results on Qwen2-VL-7B~\citep{wang2024qwen2}, a multimodal model whose vision and language modules differ substantially from LLaVA-v1.6~\citep{liu2024llavanext}.

\begin{table}[t]
    \centering
    \small
    \resizebox{\columnwidth}{!}{%
        \begin{tabular}{llll}
        \toprule
        Model &  Test &  SimRate  &  \textbf{\textcolor{black}{$\blacktriangle$}} \\
        \midrule
        \imagechain$_{C2}$  & C2  &  16.6  &  - \\
        \imagechain$_{C2-7}$  & C2  &  18.3  &  \textcolor{green}{$\blacktriangle$} 1.7 \\
        \imagechain$_{C3}$  & C3  &  16.2  &  - \\
        \imagechain$_{C2-7}$  & C3  &  21.3  &  \textcolor{green}{$\blacktriangle$} 5.1 \\
        \imagechain$_{C4-7}$  & C4--6  &  14.0  &  - \\
        \imagechain$_{C2-7}$  & C4--6  &  16.8  &  \textcolor{green}{$\blacktriangle$} 2.8 \\
        \bottomrule
        \end{tabular}
    }
    \caption{Similarity rate (SimRate) on \emph{next-scene description} when \imagechain{} is trained and evaluated on varying context lengths. \textcolor{black}{$\blacktriangle$} refers to the gains from training on C$\tau$--7 compared to the corresponding C$\tau$ trained model. \imagechain{} benefits from training on different context lengths.}
    \label{tab:cxt-lengths}
\end{table}

The results show a similar progression: once three or more prior scenes are provided, \imagechain{} dominates with +2.2\% on C4-6 and +23.1\% on C6 vs. standard fine-tuning, i.e., MLLM-FT. This underscores our argument that explicit multi-turn formulation is critical for extended temporal dependencies.
\imagechain{} on Qwen likely lags behind MLLM-FT at C2 due to its instruction overhead, which offers less value in short contexts. However, as more prior scenes become available (C3+), its multi-turn conversation design increasingly benefits from modeling temporal progression. Although both methods tie overall, reflecting the dataset’s bias toward short stories (see Figure~\ref{fig:storyframes-dist}), \imagechain{} proves more effective as context length increases.

\paragraph{Validating SimRate With Human Alignment.}
\label{sec:human-validation}
To ensure the reliability of SimRate, we conducted a human annotation study comparing Llama 3.1 70B's judgments against human evaluations on 90 diverse examples (full description and detailed analysis in Appendix~\ref{sec:appendix-humaneval}).
The Llama judge achieved 72\% accuracy (95\% CI: 62–80\%) against human majority vote, and its overall SimRate showed no significant difference from human-aggregated ratings.
While SimRate is generally aligned over all analyzed models, the Llama judge's calibration varies by model: it was conservative for \imagechain{} and MLLM-FT (underestimating their human-rated scores by 10-13 pp.) and optimistic for MLLM (+7 pp.).
Therefore, if ranked by human labels, \imagechain{}
would see its score further increase by approximately 13 points.

\paragraph{\imagechain{} Benefits from Training on Different Context Lengths.}
As shown in Table~\ref{tab:cxt-lengths}, training \imagechain{} on a diverse range of temporal spans (\imagechain$_{C2-7}$) consistently enhances performance compared to training on fixed, specific context lengths. For instance, \imagechain$_{C2-7}$ outperforms \imagechain$_{C2}$ by 1.7 pp. when tested on C2 (18.3\% vs. 16.6\%) and surpasses \imagechain$_{C3}$ by 5.1 pp. on C3 (21.3\% vs. 16.2\%). Even on longer contexts (C4-6), \imagechain$_{C2-7}$ shows a 2.8 point gain over \imagechain$_{C4-7}$ (16.8\% vs. 14.0\%). These results highlight that exposure to varied temporal spans during training enables more robust sequential reasoning and better generalization across different context lengths. Further analyses in Appendix~\ref{sec:appendix-context-length-ablation} show that when models are trained on longer contexts (C4-7), \imagechain{} demonstrates superior or highly competitive performance against baselines on both short and long evaluation contexts.

\begin{table}[t]
\centering
\small
\begin{tabularx}{\columnwidth}{l*{3}{>{\centering\arraybackslash}X}}
\toprule
Model & Comics & DL & Robo \\
\midrule
GPT-4V$^\dagger$   & \textbf{18.1} & \textbf{33.6} & 34.0 \\
Gemini$^\dagger$  & 16.3 & 21.6 & \textbf{39.4} \\
\addlinespace[0.5em]
Video-LLaMA-2$^\dagger$ & 6.8 & 20.1 & 11.2 \\
Chat-UniVi$^\dagger$ & 12.0 & 24.9 & 21.1 \\
\addlinespace[0.5em]
MLLM     & 13.3 & 23.5 & 15.2 \\
MLLM-FT  & 13.0 & 18.4 & 14.4 \\
VisualContext  & 11.2 & 18.3 & 27.8 \\
FinalScene & 13.4 & 26.5 & 19.0 \\
\imagechain & 16.2 & 20.8 & 27.1 \\
\bottomrule
\end{tabularx}
\caption{%
Out-of-domain performance (F1 score) on Comics, Daily-Life (DL), and Robotics (Robo) datasets. \imagechain{} excels in structured, event-based reasoning tasks such as comics, trailing Gemini only by 0.1.
$^\dagger$: results are taken from \citet{wang-etal-2024-mementos}.
}
\label{tab:ood-res}
\end{table}

\subsection{Out-of-Domain Performance}
Table \ref{tab:ood-res} shows \imagechain{}'s out-of-domain performance against various MLLMs and video models. Larger API-based models (GPT-4V, Gemini~\citep{team2023gemini}) are included as references, though their scale makes direct comparison challenging.

\begin{figure}[t]
    \centering
    \includegraphics[width=0.97\linewidth]{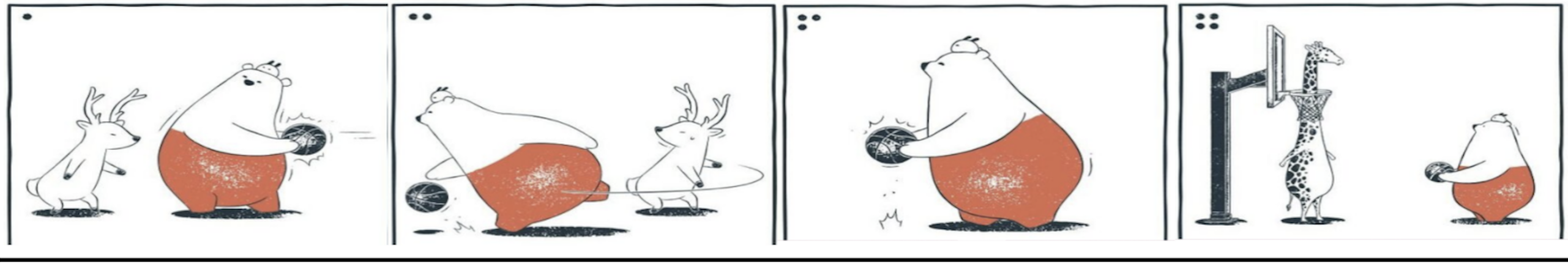}
    
    \vspace{5pt}
    \small
    \begin{tabular}{p{0.96\linewidth}}
        \textbf{MLLM-FT} \\
        \textit{In the image, a white bear is \textbf{\textcolor{orange}{standing}} on the right side while \textbf{\textcolor{orange}{holding}} a basketball in his hands and a deer is \textbf{\textcolor{orange}{standing}} on the left side and \textbf{\textcolor{orange}{look}} at the basketball.} \\ 
        \textbf{\imagechain{}} \\
        \textit{In the image, there is a white bear on the right side wearing a red lower and holding a ball in its hands while a deer on the left side is \textbf{\textcolor{MyPink}{moving}} its head. The white bear is standing and \textbf{\textcolor{MyPink}{holding}} a ball in its hands and the deer is moving its head. The white bear is \textbf{\textcolor{MyPink}{throwing}} the ball into the basketball hoop while a giraffe is \textbf{\textcolor{MyPink}{standing}} and \textbf{\textcolor{MyPink}{looking}} at the basketball hoop. The white bear is \textbf{\textcolor{MyPink}{walking}} towards the basketball hoop and \textbf{\textcolor{MyPink}{putting}} the ball into the hoop.} \\ 
    \end{tabular}

    \caption{Comparison of descriptions for sequential images in an out-of-domain setting (comics). While MLLM-FT (standard fine-tuning) tends to describe static observations, \imagechain{} captures dynamic, sequential actions. Key behavioral cues, such as action-oriented verbs, are highlighted in the text.}
    \label{fig:seq_reasoning}
\end{figure}

\textbf{Comics.}
In the comics domain, \imagechain{} trails Gemini by only 0.1 points (16.2 vs. 16.3) and surpasses all 7B baselines including video models, indicating that modeling sequential dependencies benefits structured visual narratives. Figure \ref{fig:seq_reasoning} shows how \imagechain{} captures the evolution of events over time, unlike MLLM-FT that focus on static observations. 

\textbf{Daily-Life.}
For daily life (DL) videos, \imagechain{} lags behind FinalScene (20.8 vs. 26.5 F1), suggesting that the task is closer to traditional captioning since descriptions summarize videos with fewer significant changes compared to comics, where progression between frames is more explicit.

\textbf{Robotics.}
In robotics (Robo), \imagechain{} (27.1 F1) shows a substantial improvement over MLLM (15.2 F1) and MLLM-FT (14.4 F1) as well as both video models (20.1 and 24.9 F1), highlighting that explicitly modeling image sequences improves reasoning in multi-step environments, even in out-of-domain settings. VisualContext (27.8 F1) achieves a similar performance, suggesting that leveraging prior frames without explicit text descriptions can be effective in this setting.

\section{Conclusions}
\label{sec:conclusions}
In this work, we introduced \imagechain{}, a framework designed to enhance MLLMs with explicit sequential reasoning capabilities, addressing a key limitation in existing MLLMs. Our results demonstrate substantial improvements over baseline models, with strong performance gains in both in-domain and zero-shot out-of-domain settings. %
Our findings highlight the importance of instruction-tuning within a multimodal, multi-turn conversational framework. This suggests promising directions for future work in refining temporal reasoning, scaling to more complex scenarios, and applications like video understanding or assistive robotics.

\section*{Limitations}
While \imagechain{} enhances sequential reasoning by structuring image sequences as multi-turn conversations, 
maintaining coherence across many turns may require more sophisticated techniques, which can be explored in future work. That said, our framework already models dependencies more effectively than standard MLLMs, making it a strong foundation for further improvements. 

\section*{Ethics Statement}
This work uses publicly available licensed (CC BY 4.0) datasets consistent with their intended use (research) to ensure transparency and reproducibility. While \imagechain{} enhances sequential reasoning, it may inherit biases from pre-trained models and datasets. Our framework is designed for research and development purposes, and we encourage responsible use, particularly in applications involving decision-making in sensitive domains such as healthcare and robotics. We acknowledge the use of Microsoft Copilot (\url{https://copilot.microsoft.com/}) during the development of the coding experiments. To ensure ethical data annotation, we recruited participants, balancing gender. Annotators provided informed consent and were compensated fairly at a rate of £9 per hour. Each participant annotated 12 examples (see Appendix \ref{sec:appendix-humaneval}) using a task interface designed for clarity and ease of use. The task involved no sensitive or harmful content, and all data used were synthetic or publicly available, with no personally identifiable information involved.

\section*{Acknowledgments}
The research was supported by a research grant (VIL53122) from VILLUM FONDEN, and by the European Union’s Horizon 2020 research and innovation program under grant agreement No. 101135671 (TrustLLM).

We acknowledge EuroHPC Joint Undertaking for awarding us access to Karolina, hosted by IT4Innovations, Czech Republic.

This work was supported by the Ministry of Education, Youth and Sports of the Czech Republic through the e-INFRA CZ (ID:DD-24-66).

We thank Rita Ramos for interesting discussions on this work.

\bibliography{anthology, sitcom}

\newpage
\appendix

\onecolumn
\section{Evaluation Prompt}
\label{sec:appendix-eval-prompt}
\begin{figure*}[h]
\begin{tcolorbox}[
    colback=textbox,
    colframe=headerbox,
    colbacktitle=headerbox,
    fontupper=\fontsize{8}{9}\selectfont
]
\texttt{\textcolor{cgreen}{<|start\_header\_id|>}\textcolor{cmint}{system}\textcolor{cgreen}{<|end\_header\_id|>}}\newline\newline
You are a pattern-following assistant that can only answer with "Yes" or "No". Your goal is to determine whether a predicted caption conveys a similar enough meaning to the ground truth caption provided.\texttt{\textcolor{cgreen}{<|eot\_id|>}}
\texttt{\textcolor{cgreen}{<|start\_header\_id|>}\textcolor{cmint}{user}\textcolor{cgreen}{<|end\_header\_id|>}}\newline

\#\#\# Instruction:\newline
Determine if the predicted caption conveys a similar meaning to the ground truth caption.\newline

\#\#\# Ground truth caption:\newline
A man is riding a bicycle through a park.\newline

\#\#\# Predicted caption\newline
A person is cycling along a path in a park.\newline

\#\#\# Does the predicted caption convey a similar meaning to the ground truth caption (Yes or No)?\texttt{\textcolor{cgreen}{<|eot\_id|>}} \texttt{\textcolor{cgreen}{<|start\_header\_id|>}\textcolor{cmint}{assistant}\textcolor{cgreen}{<|end\_header\_id|>}}\newline

Yes\texttt{\textcolor{cgreen}{<|eot\_id|>}}\texttt{\textcolor{cgreen}{<|start\_header\_id|>}\textcolor{cmint}{user}\textcolor{cgreen}{<|end\_header\_id|>}}\newline

Good job! Indeed, the predicted caption conveys a similar meaning to the ground truth. Both describe a person riding a bicycle in a park, even though different words are used. The core meaning is preserved.\newline

\#\#\# Instruction:\newline
Determine if the predicted caption conveys a similar meaning to the ground truth caption.\newline

\#\#\# Ground truth caption:\newline
A woman is sitting on a wooden bench in the park, reading a paperback novel under the shade of a tree.\newline

\#\#\# Predicted caption:\newline
A woman relaxes in a shaded area of the park, sitting on a bench while enjoying a book.\newline

\#\#\# Does the predicted caption convey a similar meaning to the ground truth caption (Yes or No)?\texttt{\textcolor{cgreen}{<|eot\_id|>}} \texttt{\textcolor{cgreen}{<|start\_header\_id|>}\textcolor{cmint}{assistant}\textcolor{cgreen}{<|end\_header\_id|>}}\newline

Yes\texttt{\textcolor{cgreen}{<|eot\_id|>}}\texttt{\textcolor{cgreen}{<|start\_header\_id|>} \textcolor{cmint}{user}\textcolor{cgreen}{<|end\_header\_id|>}}\newline

Great! Although the wording differs, the predicted caption captures the essence of the ground truth. Both describe a woman sitting on a bench in a shaded park area, reading a book. While the predicted caption simplifies certain details, such as omitting the specific mention of the "paperback novel" and "under the shade of a tree," it still conveys the same overall scene and activity, making the meaning similar.\newline

Let's do one more. Remember to answer with one word either "Yes" or "No".\newline

\#\#\# Instruction:\newline
Determine if the predicted caption conveys a similar meaning to the ground truth caption.\newline

\#\#\# Ground truth caption:\newline
\textcolor{corange}{\texttt{[$\ldots$]}}\newline

\#\#\# Predicted caption:\newline
\textcolor{corange}{\texttt{[$\ldots$]}}\newline

\#\#\# Does the predicted caption convey a similar meaning to the ground truth caption (Yes or No)?: \texttt{\textcolor{cgreen}{<|eot\_id|>}}
\texttt{\textcolor{cgreen}{<|start\_header\_id|>} \textcolor{cmint}{assistant} \textcolor{cgreen}{<|end\_header\_id|>}}

\end{tcolorbox}
\caption{%
Evaluation prompt used with Llama 3.1 70B to annotate the predicted descriptions. The prompt structure is adapted towards description prediction from \href{https://github.com/tatsu-lab/alpaca_eval/blob/main/src/alpaca_eval/evaluators_configs/alpaca_farm/chatml_b1_chat_v0_without_inputs.txt}{Alpaca-Eval}. The positions are indicated by placeholders \textcolor{corange}{[$\ldots$]}, where the ground truth and predictions to be annotated are inserted.
}
\end{figure*}

\twocolumn

\section{Human Annotation Study}
\label{sec:appendix-humaneval}

\paragraph{Study Design.}

\begin{figure}[h!]
    \centering
    \includegraphics[width=0.95\linewidth]{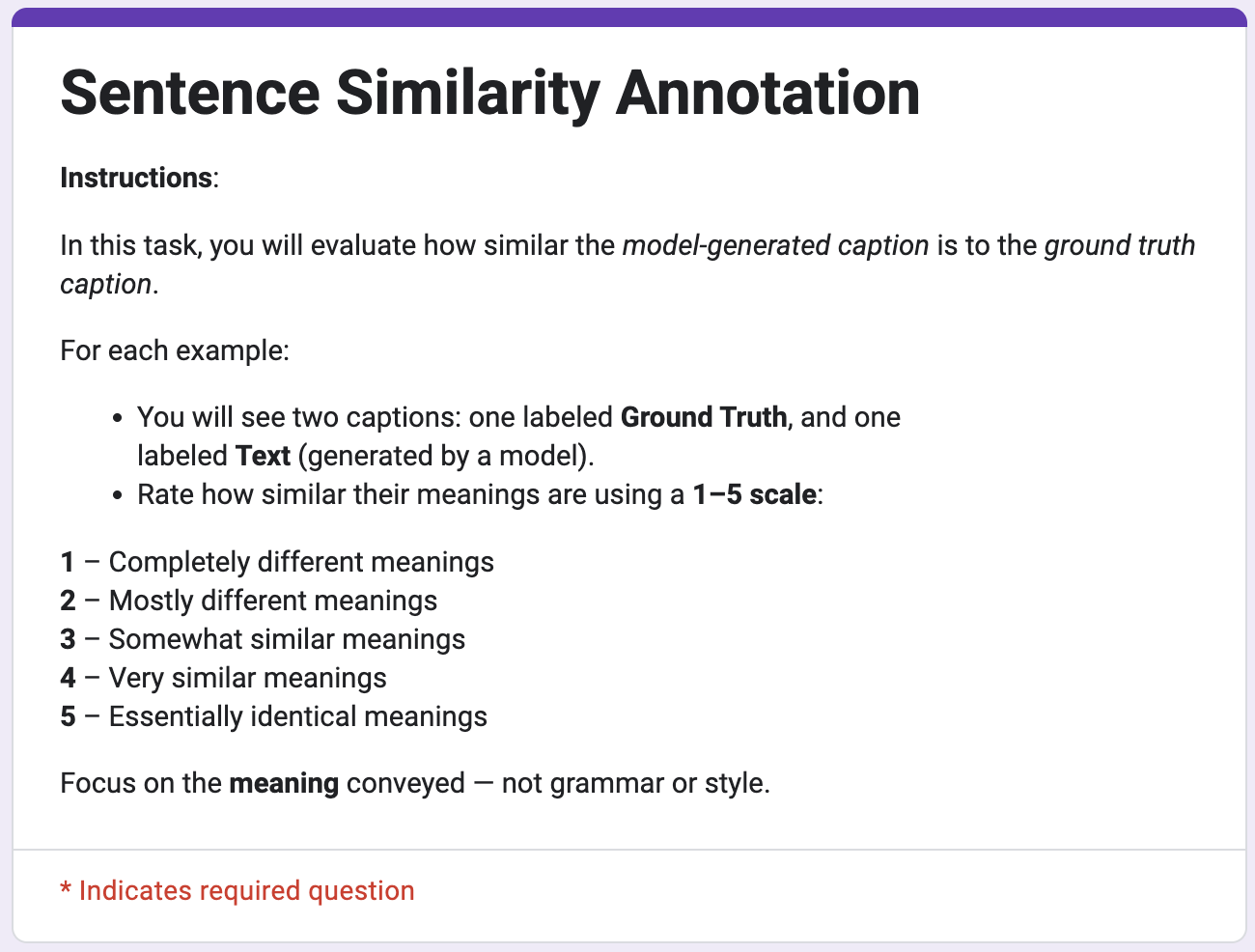}
    \vspace{-0.2cm}
    \caption{Instructions of human annotation study}
    \label{fig:annotation-instructions}
\end{figure}

\begin{figure}[h!]
    \centering
    \includegraphics[width=0.95\linewidth]{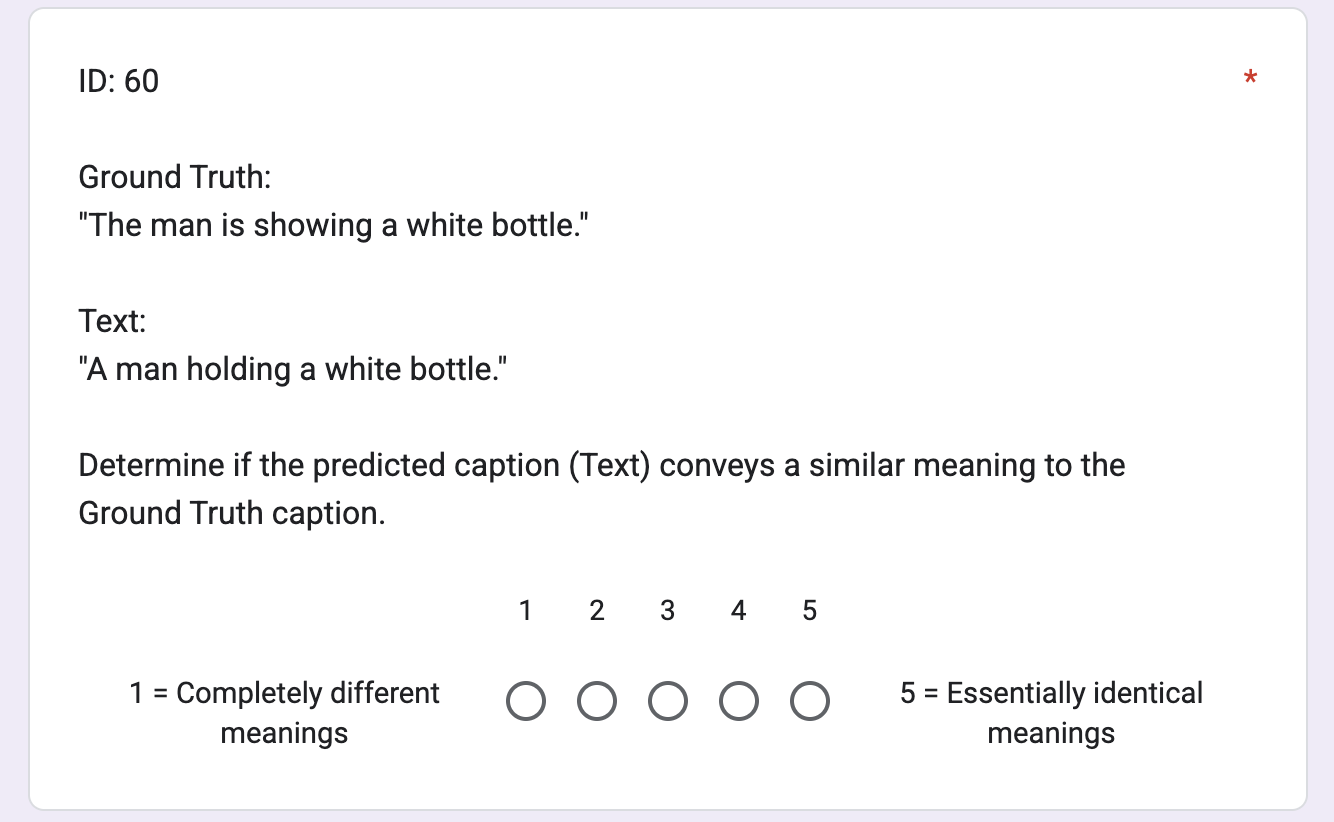}
    \vspace{-0.2cm}
    \caption{Example of the annotation interface}
    \label{fig:annotation}
\end{figure}

Ninety examples were selected for evaluation, stratified by context length and sampled proportionally from the outputs of \imagechain{}, MLLM, and MLLM-FT. Three human annotators were recruited via Prolific (\url{https://www.prolific.com/}), ensuring a balanced gender distribution and a minimum education level of an undergraduate degree. Annotators provided informed consent and were compensated at a rate of £9 per hour, in line with fair pay guidelines. Each participant annotated 12 examples, which included three gold-standard control examples (labeled by the authors for quality control; annotators who failed these controls were excluded from the final analysis).
For each example, annotators compared the ground truth next-scene description with a model-generated description, rating the semantic overlap on a five-point Likert scale (1 = ``completely different meanings'', 5 = ``essentially identical meanings''). The instructions and the annotation interface are shown in Figure~\ref{fig:annotation-instructions} and Figure~\ref{fig:annotation} respectively.

\paragraph{Data Analysis and Inter-Annotator Agreement.}
For subsequent quantitative analysis~\citep{hallgren2012computing}, the 5-point Likert scores were binarized: scores $\geq$ 3 were considered as positive class. This threshold resulted in a positive class prevalence of 23\% in the human-annotated dataset.
The reference label for each example was determined by a majority vote among the three annotators.

Inter-annotator agreement is fair~\citep{landis1977_measurement-8d20e0b8-89d8-3d65-bcf5-8c19d56ec4ab}, with Fleiss' $\kappa$ = 0.30~\citep{fleiss1971_measuring-CTX:C6682} and Krippendorff's $\alpha_{\text{ordinal}} = 0.32$~\citep{krippendorff2018content} when computed on the original 1-to-5 ranks, indicating moderate consistency in how annotators judge semantic similarity, even if they differ on the exact rating level.

\paragraph{Alignment with Llama Judge.}
Compared to human annotations, the Llama 3.1 70B judge achieves 72\% accuracy, Wilson~\citep{wilson1927probable} 95\% CI: 62–80\%, with balanced errors (10 false positives, 15 false negatives).
McNemar's exact test for directional bias~\citep{mcnemar1947note} produced a p-value of 0.42, indicating no significant systematic bias in one direction over the other.
Overall, the SimRate produced by the Llama 3.1 70B judge differed from the human-aggregated SimRate by 5.6 percentage points (95\% CI for the difference: –17.3 pp to +6.2 pp).
Since this confidence interval includes zero, there is no statistically significant difference between the overall SimRate produced by the Llama judge and human evaluators.

\begin{figure*}[t!]
    \centering
    \includegraphics[width=0.97\linewidth]{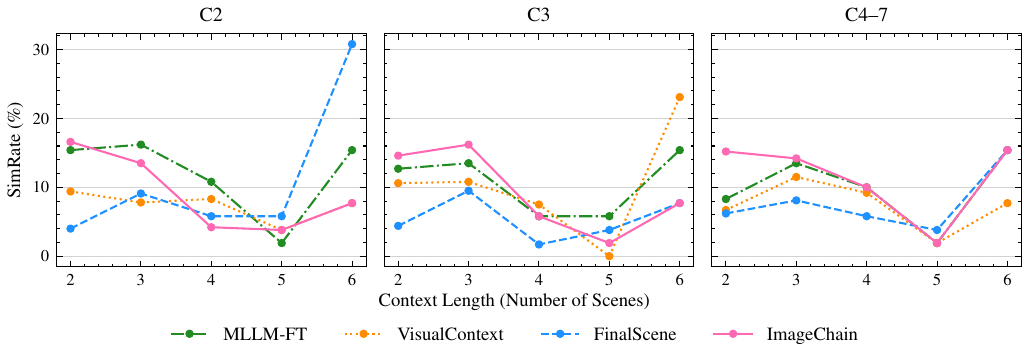}
    \caption{Model performance (SimRate) for fine-tuned models trained on different context lengths (C2, C3, C4-7). \imagechain{} achieves the highest overall SimRate when trained on long contexts (C4-7).}
    \label{fig:sub-charts}
\end{figure*}

\paragraph{Model-Specific Calibration Differences.}
While the overall alignment was accurate, the Llama 3.1 70B judge exhibited some model-specific calibration differences when compared to human ratings:
\begin{itemize}
\item \textbf{For \imagechain{} outputs:} The Llama judge was conservative. It achieved 87\% accuracy on these samples but tended to underestimate positive cases. The human-rated SimRate for \imagechain{} was 13 percentage points higher than the Llama-judged SimRate.
\item \textbf{For MLLM outputs:} The Llama judge was slightly liberal. It achieved 67\% accuracy and tended to overestimate positive cases. The human-rated SimRate for MLLM was 7 percentage points lower.
\item \textbf{For MLLM-FT outputs:} The Llama judge was again conservative. It achieved 67\% accuracy, and the human-rated SimRate for MLLM-FT was 10 percentage points higher.
\end{itemize}

These differences suggest that if the main results were ranked purely based on human labels, \imagechain{} and MLLM-FT would likely see their scores increase relative to the Llama-judged SimRates, while MLLM's score would decrease.

\section{Training and Implementation Details}
\label{sec:appendix-implementation_details}

We fine-tune both LLaVA-v1.6~\citep{liu2024llavanext} and Qwen2-VL-7B~\citep{wang2024qwen2} using low-rank adaptation~\citep[LoRA]{hu2021lora} in combination with DeepSpeed ZeRO-3, running approximately 1-2 hours per model on 8 NVIDIA A100 GPUs.
The baseline training is conducted over 3 epochs on the StoryFrames dataset with a per-device batch size of 4 and gradient accumulation of 1.
LoRA is configured with a rank \(r=128\) and \(\alpha=256\).
A learning rate of \(2 \times 10^{-5}\) is applied, with a cosine learning rate schedule and a warmup ratio of 3\%.
For optimization, the adaptive momentum optimizer with decoupled weight decay~\citep{loshchilov2019adamw} is used.
For \imagechain{} trained on C2 and C3 we adjust the number of training epochs to 5, and to 7 epochs when trained on C4--7 due to the smaller training sample sizes.
All other training settings remain unchanged.

\section{Context Length Ablation}
\label{sec:appendix-context-length-ablation}

Figure \ref{fig:sub-charts} shows the models performance for fine-tuned models trained on different context lengths. \imagechain{} achieves the highest overall SimRate when evaluated on C2-6, particularly when trained on longer contexts, reaching 13.6\% when trained on C4-7. This surpasses other models, such as MLLM-FT, which achieves 9.8\% under the same training conditions. MLLM-FT excels in short contexts but struggles with longer dependencies, suggesting limitations in handling extended sequences without explicit sequence modeling. VisualContext under-performs on longer sequences (7.7\% trained on C4-7 and evaluated on C6), highlighting the benefit of including text descriptions for fine-tuning in long contexts.

\end{document}